\documentclass[letterpaper,10pt,conference]{ieeeconf}  

\IEEEoverridecommandlockouts                              

\usepackage{booktabs}
\usepackage{url}

\usepackage{amsmath,amsthm,amsfonts,amssymb}
\usepackage{bbold}
\usepackage{xcolor}
\usepackage[english]{babel}
\usepackage[pdftex]{graphicx}
\usepackage{subcaption}
\usepackage{algorithm}
\usepackage{comment}
\usepackage{float}
\usepackage[hidelinks]{hyperref}
\usepackage{cite}
\usepackage[noend]{algpseudocode}

\theoremstyle{definition}
\newtheorem{definition}{Definition}[]

\theoremstyle{assumption}
\newtheorem{assumption}{Assumption}[]

\newtheorem*{remark}{Remark}

\newcommand{\rev}[1]{{#1}}

\newcommand{\etal}{\textit{et al.}}

\renewcommand{\baselinestretch}{0.9255}

\title{\LARGE \bf
Chemistry-Inspired Pattern Formation with Robotic Swarms 
}

\author{Paulo Rezeck and Luiz Chaimowicz
\thanks{Paulo Rezeck and Luiz Chaimowicz are with the Department of Computer Science, Universidade Federal de Minas Gerais,
        Brazil. 
        Email: {\tt\small \{rezeck,chaimo\}@dcc.ufmg.br}.
        This work was partially supported by CAPES, CNPq, and Fapemig.}%
}

\begin{document}

\maketitle
\thispagestyle{empty}
\pagestyle{empty}


\begin{abstract} 

Self-organized emergent patterns can be widely seen in particle interactions producing complex structures such as chemical elements and molecules. Inspired by these interactions, this work presents a novel stochastic approach that allows a swarm of heterogeneous robots to create emergent patterns in a completely decentralized fashion and relying only on local information. Our approach consists of modeling the swarm configuration as a dynamic {\em Gibbs Random Field} (GRF) and setting constraints on the neighborhood system inspired by chemistry rules that dictate binding polarity between particles. Using the GRF model, we determine velocities for each robot, resulting in behaviors that lead to the \rev{creation} of patterns or shapes. 
Simulated experiments show the versatility of the approach in producing a variety of patterns, and experiments with a group of physical robots show the feasibility in potential applications.

\end{abstract}

\section{Introduction}
\label{sec:intro}

Systems of a large number of particles dynamically interacting pairwise produce extraordinarily complex patterns~\cite{saintillan2008instabilities,von2012predicting}. Well-known examples of patterns generated by these systems are molecular structures that emerge from atomic interactions depending on environmental conditions~\cite{verwey1947theory, gillespie1957inorganic}. The study of such systems pervades many disciplines ranging from physics and chemistry to biology and pharmacy, having high societal and economic impact. In particular, one may use it to design new compounds or materials and understand biological systems.

Although challenging, the study of such systems may provide powerful tools for a wide variety of applications in robotics, especially for swarm robotics in pattern (shape) formation problems. The pattern formation problem may be defined as the coordination of a group of robots to get into and maintain a formation with a certain shape~\cite{bahceci2003review}. A key aspect for the applicability of these models in swarm robotics is the requirement for distributed and decentralized processing relying only on local information. Models with these characteristics bring several practical advantages allowing scalability, resiliency, and adaptability. Examples of potential applications would be oil spill containment or cleaning in oil plants~\cite{kim2012toward,shah2018autonomous} and constructing structures such as a temporary bridge that could dynamically adjust its size and shape to fit different environmental conditions~\cite{rong2020employing}.

\begin{figure}[!t]
		\centering
        \includegraphics[width=.99\columnwidth]{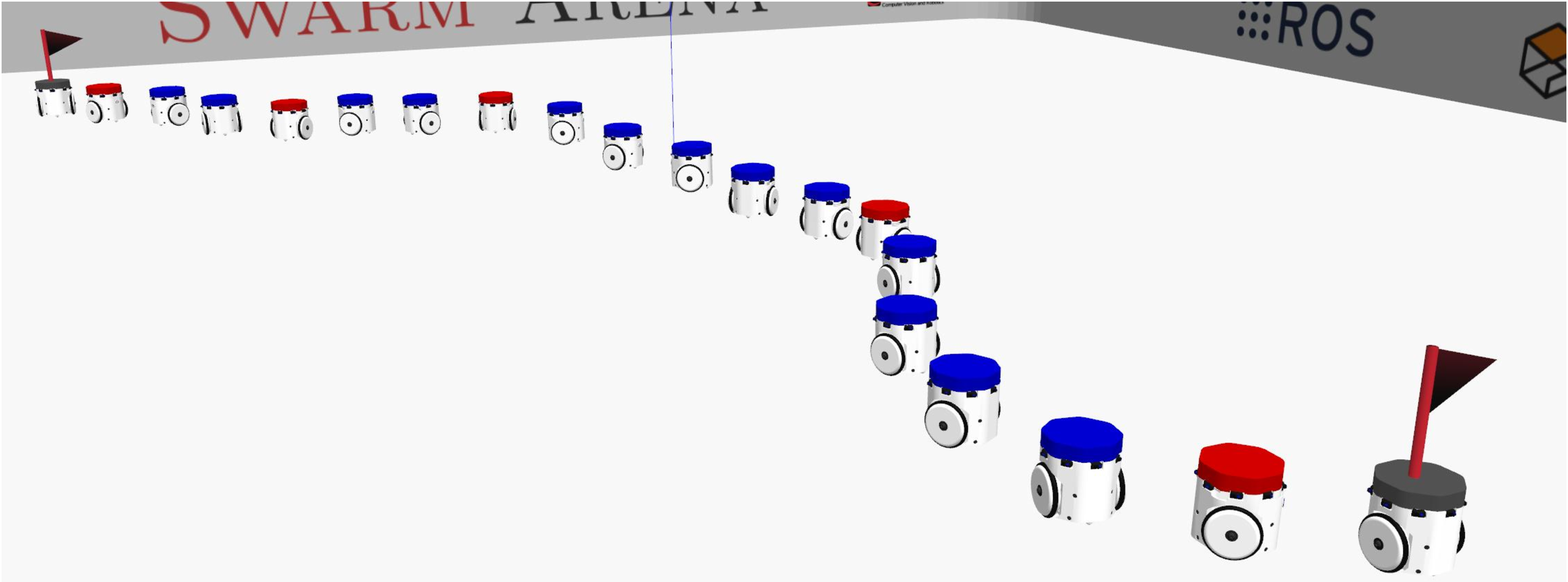}
        \caption{Robots mimicking atoms of \textit{carbon} (red) and \textit{oxygen} (blue) create emergent chain patterns useful for dynamic bridge-building applications. Robots with red flags indicate the beginning and end of the chain.} 
	\label{fig:patternformation}
\end{figure}

By revisiting some of the concepts and theories applied in particle interactions and molecular structures formation, we create a simplified model suitable for robot swarms. This work presents a novel stochastic and decentralized approach that allows a swarm of heterogeneous robots to emerge with interesting patterns relying entirely on local interactions with neighbors. 
Our approach consists of modeling the robot swarm as a dynamic Gibbs Random Field (GRF) and defining 
the neighborhood system inspired by the Octet rule used in chemistry. 
By setting the GRF's potential energy as a combination of Coulomb-Buckingham potential and kinetic energy, the robots can safely navigate through a bounded environment and bind with others forming global patterns using only local interactions. Figure~\ref{fig:patternformation} shows a swarm \rev{forming a chain} in the environment by using robots that mimic atoms of \textit{carbon} and \textit{oxygen}.

The remainder of this paper is organized as follows: we review and discuss some of the most relevant works in swarm robotics regarding pattern and shape formation in Section~\ref{sec:relatedwork}. 
Our methodology is detailed in Section~\ref{sec:methodology}, and experimental results in simulated and real scenarios are presented in Section~\ref{sec:experiments}. Finally, we close this paper with our conclusions and directions for future work in Section~\ref{sec:conclusion}.

\section{Related Work}
\label{sec:relatedwork}

Pattern formation occurs in nature at all scales and is a fundamental question across interdisciplinary research, including topics on physical chemistry~\cite{lopez2018pattern, zhang2020pattern}, 
cosmology~\cite{liddle2000cosmological}, and biochemistry~\cite{maini1997spatial, kai2019formation}. Several patterns or shape formation approaches have also been proposed in the literature for multi-robot systems~\cite{bahceci2003review, varghese2009review, liu2016survey, oh2017bio, chennareddy2017modular}. Most of these works assume global information, which allows each robot in the swarm to directly perceive every other robot~\cite{belta2002trajectory,egerstedt2001formation,pereira2008adaptive,vickery2021mean}. Such assumptions allow for fast and efficient convergence of the swarm in pattern formation but may be unrealistic in real applications. Others assume that global information is not always available and deal with the task allocation problem in which robots must coordinate to reach different predefined positions that form patterns or shapes~\cite{hsieh2008decentralized,varghese2009towards,rahmani2009controllability, alonso2011multi,wang2020shape}. A typical strategy to avoid the requirement of defining positions describing the patterns consists of using seed robots in which some robots do not move and act as a reference to the others helping the swarm to create complex global patterns~\cite{grushin2010parsimonious, rubenstein2014programmable}. Differently from these, our approach assumes minimal and local-only information to produce interesting patterns that resemble molecular shapes. \rev{Minimalist approaches are attractive for swarm robotics due to the low sensing capability of the robots, and recent work has shown their feasibility for self-organization problems~\cite{gauci2014self,st2018circle,lavergne2019group,mitrano2019minimalistic}.}
In the remaining of this section, we discuss some works that tackle the problem using mainly local information and then present the main contributions of this work. 



Sahin~\etal~\cite{sahin2002swarm} designed a robotic system called swarm-bots. The robots can connect to or disconnect from each other using a grasping mechanism enabling self-assemble into different kinds of structures. Inspired from social insect studies~\cite{camazine2020self}, the authors employed a probabilistic approach to control the robots. Preliminary results in simulation show that the robot can create patterns, such as a single stripe pattern, beyond the perceptions of individual robots. Despite having a complex dynamic, some aspects of the resulting patterns, such as the mean length of chains, can be controlled through parameters such as the disconnection probability in chain formation behavior. Although the authors do not demonstrate the pattern formation using real robots, the robotic system and control strategy favored the development of several other studies, such as aggregation~\cite{bahgecci2005evolving} and self-assembly~\cite{dorigo2004evolving}.



Slavkov~\etal~\cite{slavkov2018morphogenesis} proposed a morphogenesis approach inspired by spontaneous phenomena observed in some biological systems during embryogenesis~\cite{economou2012periodic}. The shapes emerge in a fully self-organized way. The robots rely only on local interactions with neighbors and do not require maps, coordinate systems, or preprogrammed seed robots. The approach uses the concept of robot migration (in analogy with natural developmental biology) and gene regulatory networks (GRN) to create a self-organizing Turing process for pattern formation. The authors successfully demonstrated their method in a swarm of 300 real robots, showing robustness and adaptability in forming Turing patterns.

Further, Carrillo-Zapata~\etal~\cite{carrillo2019toward} extended the previous approach to increase the controllability of the system, enabling the formation of specific patterns. The author designed a morphogenesis algorithm based on local gradients for swarms of simple and noisy robots capable of communicating among them. By setting three parameters, robots self-organize to grow controllable shapes while maintaining the communication network. Results demonstrated that the swarm emerges with the rich morphospace of quantitatively different shapes by changing these parameters.


Li~\etal~\cite{li2019decentralized} proposed a case study of pattern formation that can be applied to any shape described as a 2D point cloud. To achieve this, the authors present an algorithm that transforms a given point cloud into an acyclic directed graph shared among the swarm members. This graph is used by the control law to allow a swarm of robots to progressively form the target shape based only on local decisions. This means that free robots (i.e., not yet part of the formation) find their location based on the perceived location of the robots already in the formation. Extensive simulations and experiments on real robots show the effects of swarm size. Results indicate that the algorithm is robust to noise and can handle different formations and shapes. 

Coppola~\etal~\cite{coppola2019provable} presented a \rev{minimalistic approach} to generate a local behavior that allows a swarm of \rev{homogeneous} robots to self-organize into a desired global pattern by relying only on the relative location of their closest neighbors. The generated local behavior is a probabilistic local state-action map, and robots follow policies to select appropriate actions based on their current perception of their neighborhood. Simulations showed the method's robustness against sensor noise and demonstrated the formation of patterns using micro air vehicles. In addition, the authors discuss the scalability of the method and synchronization issues between robots. \rev{Although the method uses robots with limited sensory apparatus, it requires a connected topology in the initial configuration of the swarm and an environment discretized by lattices to guarantee the convergence of the swarm in the pattern without suffering from deadlocks.} 

Unlike other works, we took inspiration from the Octet rule in chemistry to generate patterns with a robotic swarm. Basically, the Octet rule defines the number of bonds each atom preferably makes. The application of the Octet rule in the robotics's context is yet restricted. Shiu~\etal~\cite{shiu2010modular} presented the design of modular robots that uses the Octet rule to dictate attraction force and motion capability. \rev{Randall~\etal~\cite{randall2016simulating} proposed a decentralized mechanism that aims to simulate chemical reactions using a swarm of miniature robots. The motivation for this work lies in the development of an educational tool to simulate chemical reactions capturing either behavioral or embodied aspects, differently from other tools such as computer simulations or ball-and-stick models. The proposed mechanism replicates what would be expected by simulations of physical-chemical models as faithfully as possible. It uses various built-in sensors to detect neighboring robots and dictate the bonding rules by using direct communication between robots that periodically broadcast state messages.}

\rev{Differently from these works, our approach takes the Octet rule as inspiration to create a mechanism for pattern formation. In addition, the method presented here brings other functionalities such as obstacle avoidance, cohesive navigation, and binding with other robots using only local interactions.}
Our method consists of modeling the swarm as a GRF and constraining the neighborhood system by the Octet rule. This allows us to formalize a probability function that indicates, in a decentralized way, which velocity is most likely for the robot given only the neighborhood information. Moreover, \rev{it} 
does not rely on global information, goal assignments, communication topology, or preprogrammed seeds to produce patterns. By setting appropriate potential functions and neighborhood constraints, the swarm forms different patterns resembling molecular structures.

\section{Methodology}
\label{sec:methodology}

The proposed methodology extends our previous work~\cite{rezeck2021flocking}, in which we designed and evaluated a novel approach that allowed a swarm of heterogeneous robots to achieve flocking and segregative behaviors simultaneously. After further exploring and improving our method, we realized that by incorporating dynamic constraints to the neighborhood system and adequately defining the swarm heterogeneity and GRF's potentials, the swarm would be able to produce specific patterns, a more complex and restrictive task in comparison to flocking, and with more tangible applications. As mentioned, by taking inspiration from the Octet rule in chemistry, 
we propose a computational-efficient mechanism to constrain the neighborhood system used by the Gibbs Random Field (GRF), enabling the formation of patterns that resemble molecular structures. A detailed description of our method is given in the next sections.

\subsection{Formalization}
Given a set $\mathcal{R}$ of $\eta$ heterogeneous robots moving in a bounded region within the two-dimensional Euclidean space\footnote{We assume two-dimensional space for convenience but one can straightforwardly extend it to a three-dimensional space.}, the objective is to find an appropriate velocity for each robot in a decentralized way that leads the entire swarm to converge towards the global minimum of the potential. Before we detail our method, let us define some concepts and assumptions.

\begin{definition}[Robot state]
The state of the $i$-th robot at time step $t$ is represented by its pose $\mathbf{q}_i ^{(t)}$ and velocity $\displaystyle{\dot{\mathbf{q}_i}^{(t)} = \mathbf{v}_i^{(t)}}$, which is bounded by $v_{max}$, $||\mathbf{v}_i^{(t)}|| \leq v_{max}$. 
\end{definition}

\begin{definition}[Heterogeneity]
The swarm heterogeneity is modeled by a partition $\tau = \{{\tau}_1, ..., {\tau}_u \}$, $u \leq \eta$, with each ${\tau}_k \subset \mathcal{R}$ containing exclusively all robots of type $k$, {\it i.e.} $\forall (j,k) : j \neq k \ \rightarrow {\tau}_k \cap {\tau}_j = \emptyset$. The heterogeneity is defined by the mass $m$ and electrical charge\footnote{The electrical charge is just a parameter and it has no tangible concept in this context of swarm robotics.} $c$ of the robot so that,  $\displaystyle{\forall (i,j) \in \tau_k : \tau_k \subset \tau \rightarrow m_i \triangleq m_j}$ and $c_i \triangleq c_j$.
\end{definition}


\begin{assumption}[Motion model]
For convenience, we assume the robots are driven by a holonomic kinematic model\footnote{This assumption can be relaxed as will be shown in Section \ref{Sec:RealExperiments}.} with a known motion model represented by $\displaystyle{\mathcal{K} : (\mathbf{q}_i^{(t)}, \mathbf{v}_i^{(t)}) \rightarrow (\mathbf{q}_i^{(t+1)})}$. 
\end{assumption}

\begin{assumption}[Sensing]
The robots have a circular sensing range of radius $\lambda$. Within this radius the robot can estimate the relative position and velocity of other robots as well as their type, and also obstacles in the environment. 
\end{assumption}


\begin{definition}[Neighborhood system]
\label{def:neighborhood}
The neighborhood system for the $i$-th robot, constrained by the sensing range $\lambda$, defines a set of robots: 
\begin{equation*}
    \mathcal{N}_i \triangleq \{j \in \mathcal{R}: j \neq i, ||\mathbf{q}_j - \mathbf{q}_i|| \leq \lambda \},
    \label{eq:neighborhood}
\end{equation*}
where $||\mathbf{q}_j - \mathbf{q}_i||$ is the Euclidean norm between the $i$-th and $j$-th robots.
\end{definition}

Definition~\ref{def:neighborhood} states that the neighborhood system is restricted only by the sensing range $\lambda$ and do not differentiate robots of different types. By considering electrical charge as heterogeneity parameters, one may incorporate constraints that allow robots to experience different levels of interaction with their neighbors.

\subsection{Octet rule constrained neighborhood}
Inspired by some concepts and models of atom interactions in chemistry, we propose a mechanism to restrict a robot to only bind with a certain number of other robots of specific types within its neighborhood. This mechanism is motivated by the \textit{Octet rule}~\cite{gillespie2002octet}, a relatively simple rule which uses an electron counting formalism for predicting bonding. Based on the Octet theory, the Octet rule generally states that atoms tend to combine so that each of them has eight electrons in their valence shell. The principle is that molecules tend to be more stable when the outer electron shell of each of their atoms is filled with eight electrons. In fact, in nature, all systems tend to acquire as much stability as possible. For example, atoms bind together to form molecules to increase their stability.


In the context of heterogeneous swarm robots, the proposed mechanism inspired by the \textit{Octet rule} works as a constraint to describe how the robots should interact. Depending on their state and neighborhood, they may decide to stay closer to one type of robot and away from other types.


From a mathematical point of view, one may create such mechanism by ordering the robots by their electrical charges and relative distances in a two-dimensional data structure. 
Formally, let us define the concept of ordered neighborhood concerning the $i$-th robot.

\begin{definition}[Ordered neighborhood]
\label{def:orderedneighbornood}
The ordered neighborhood for the $i$-th robot concerning the Euclidean distance is defined as an ordered set $\displaystyle{\bar{\mathcal{N}}_i \triangleq  (\mathcal{N}_i, \preceq)}$, where $\forall (j,k) \in \bar{\mathcal{N}}_i \rightarrow ||\mathbf{q}_j - \mathbf{q}_i|| \leq ||\mathbf{q}_k - \mathbf{q}_i||$ and $(j,k)$ are ordered pairs.
\end{definition}

Once the neighbors of the $i$-th robot are ordered by distance, let us define the concept of bond partition used to group robots of the same type.

\begin{definition}[Bond partition]
\label{def:bondpart}
The bond partition for the $i$-th robot is a ordered partition $\bar{\mathcal{B}}_i  \triangleq  (\mathcal{B}_{i,1}, ..., \mathcal{B}_{i,u}; \succeq)$, where the ordered set $(\mathcal{B}_{i,p})$ is a ordered neighborhood containing robots that have the same electrical charge among them, $\forall(j, k) \in (\mathcal{B}_{i,p}) \rightarrow ||c_j|| = ||c_k||$. Robots with higher charges have precedence over the ones with lower charges, {\it i.e.} $\forall j \in (\mathcal{B}_{i,p})$ and $\forall k \in (\mathcal{B}_{i,p+1}) \rightarrow ||c_j|| > ||c_k||$, where the pair $(p, p+1)$ define two consecutive ordered sets  in $\bar{\mathcal{B}}_i$. 
\end{definition}



From the previous definition, we can establish a data structure that allows ordering the robots within the neighborhood both by their distance and their electrical load.
Now, let us state constraints restricting the number of neighbors the robot can bind. 

\begin{definition}[Maximum bond constraints]
\label{def:bondconstrains}
The bond partition $\bar{\mathcal{B}}_i$ for $i$-th robot has a limited number of robots defined by $\mathcal{B}^\text{max}_i$, $|\bar{\mathcal{B}}_{i}| \leq \mathcal{B}^\text{max}_i$, and robots of the same charge have the same limit, $\displaystyle{\forall(i,j) \in \tau_k : c_i = c_j \rightarrow \mathcal{B}^\text{max}_i \triangleq \mathcal{B}^\text{max}_j}$. Moreover, the number of robots in each subset of $\bar{\mathcal{B}}_i$ is also restricted by $|\bar{\mathcal{B}}_{i,p}| \leq \mathcal{B}^\text{max}_{i,p}$.
\end{definition}


After defining the bond partition, we create a procedure that reduces neighboring robots to such data structure. The mechanism is presented in  Algorithm~\ref{alg:bondpartition}. In general, the mechanism prioritizes bonds with closer robots with higher electrical charges, respecting the maximum bond constraints.

\begin{algorithm}[t]
\caption{Generating $\mathcal{B}_i$ for $i$-th robot}\label{alg:bondpartition}
\begin{algorithmic}[1]
\Procedure{BondPartition}{$\mathcal{N}_i$}
    \State $\mathcal{B}_i \gets \emptyset$ \Comment{Creating bond partition}
    \State $\bar{\mathcal{N}}_i \gets \textbf{sort}(\mathcal{N}_i)$ \Comment{Sorting neighbors by distance}
    \For{$j \in \bar{\mathcal{N}_i}$}
        \If{$|\mathcal{B}_{j,c_i}| < \mathcal{B}^\text{max}_{j,c_i})$} \Comment{Add robot $j$-th robot}
        \If{$|\mathcal{B}_{i, c_j}| < \mathcal{B}^{\text{max}}_{i,c_j}$ } 
            \State $\mathcal{B}_{i,c_j} \gets \mathcal{B}_{i,c_j} + j$ 
        \EndIf
        \EndIf
    \EndFor
    \State $\bar{\mathcal{B}}_i \gets \textbf{sort}(\mathcal{B}_i)$ \Comment{Sorting partition by charge value}
    \State $\text{bonds} \gets 0$ \Comment{Count the number of bond}
    \For{$c_k \in \bar{\mathcal{B}}_i$}
        \For{$j \in \mathcal{B}_{i, c_k}$}
            \If{$\text{bonds} < \mathcal{B}^\text{max}_i$ } 
                \State $\text{bonds} \gets \text{bonds} + 1$ \Comment{Add bond}
            \Else
                \State $\mathcal{B}_{i, c_k} \gets \mathcal{B}_{i, c_k} - j$ \Comment{Remove $j$-th robot}
            \EndIf
        \EndFor
    \EndFor
    \State \textbf{return} $\bar{\mathcal{B}}_i$
\EndProcedure
\end{algorithmic}
\end{algorithm}

\begin{remark}
\label{def:bondbehavior}
The swarm \rev{generates} different patterns by setting different maximum bond constraints $\mathcal{B}^\text{max}_{i,p}$ for each $\mathcal{B}_{i,p} \subset \mathcal{B}_i$ and respecting the previous definitions.
\end{remark}

\begin{figure}[t]
		\centering
		\includegraphics[width=.75\columnwidth]{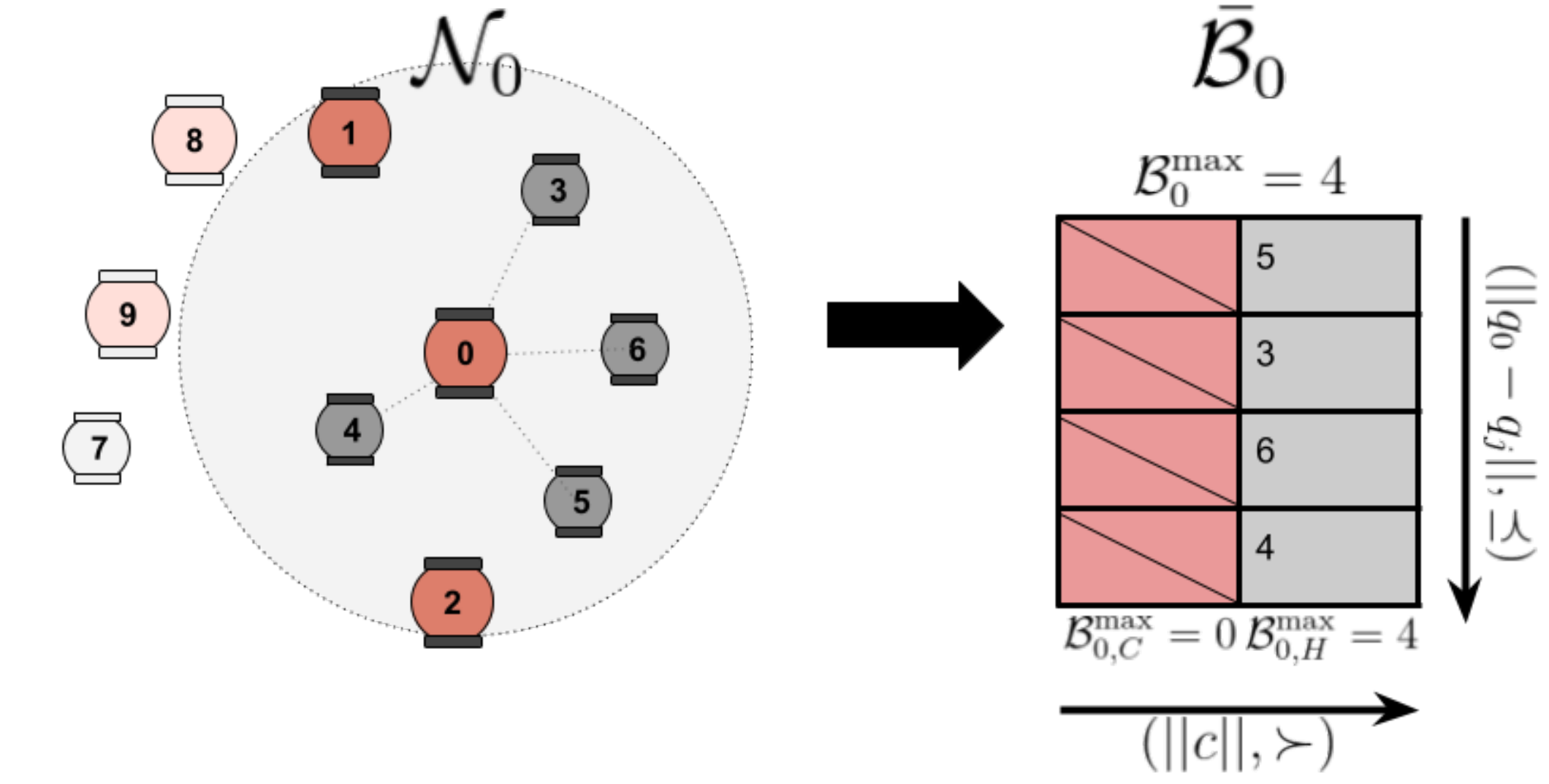}
	\caption{State of the bond partition for a given robot and its neighborhood. \textit{Carbon}-like robots (red) can only bond with a maximum of four \textit{hydrogen}-like (gray) robots.} \vspace{-0.5cm}
	\label{fig:octerneighborhood}
\end{figure}

Such mechanism forces robots to interact differently depending on the maximum bond constraints. For example, let us assume a swarm composed of two types of robots that resemble the \textit{carbon} and \textit{hydrogen} atoms. The \textit{carbon}-like robots can bind a maximum of $\mathcal{B}^{\text{max}}_{C} \triangleq 4$~robots in their neighborhood, and the \textit{hydrogen}-like robots can only bind with $\mathcal{B}^{\text{max}}_{H} \triangleq 1$~robot. 
By setting the maximum number of robots of type \textit{carbon} and \textit{hydrogen} each type can bind to, the swarm will produce different patterns. Suppose we define that \textit{carbon}-like robots can bind with $(\mathcal{B}^{\text{max}}_{C, C}, \mathcal{B}^{\text{max}}_{C, H}) \triangleq (0, 4)$ and the \textit{hydrogen}-like robots with $(\mathcal{B}^{\text{max}}_{H, C}, \mathcal{B}^{\text{max}}_{H, H}) \triangleq (1, 0)$. In that case, we should observe the formation of a structure resembling the \textit{methane} molecule. Figure~\ref{fig:octerneighborhood} shows the state of the bond partition of a \textit{carbon}-like robot and its neighborhood.


After defining a new neighborhood structure, we proceed with the modeling of the swarm configuration as a dynamic GRF and how the velocities of each robot are set so that the patterns emerge.

\subsection{Applying GRFs concepts to swarm robotics}
\label{subsec:grf_extension}
We model and control the swarm of robots as a dynamic \textit{Gibbs Random Fields} (GRFs). 
Due to space restrictions, here we only present the key aspects of this approach without providing a thorough background. More details can be found in~\cite{rezeck2021flocking}.



Formally, a random field is an undirected graph $\displaystyle{\mathbf{G} = (\mathcal{R},\mathbf{E})}$ with each vertex representing one of the robots. The neighborhood system presented previously induces the configuration of the graph $\mathbf{G}$ by establishing an edge between two robots if they are bound. A random field on $\mathbf{G}$ is a collection of random variables $\mathbf{X} = \{X_i\}_{i \in \mathcal{R}}$ and, for each $i \in \mathcal{R}$, let $\Lambda_i$ be a finite set called the phase space for the $i$-th robot. A configuration of the system $\mathbf{X}$ at time step $t$ is defined as $\displaystyle{\mathbf{x}^{(t)} = \{\mathbf{v}_1,... ,\mathbf{v}_\eta\}}$, where $\mathbf{v}_i \in \Lambda_i$ and represents the velocities performed by each robot at that time step.

The random field $\mathbf{X}$ is defined as a GRF if the joint probability density of the system is represented by,
\begin{equation}
    P(\mathbf{X} = \mathbf{x}) = \frac{1}{Z}e^{-\frac{H(x)}{T}}, \textrm{ with } Z = \sum \limits_{z} e^{-\frac{H(z)}{T}},
    \label{eq:gibbsdistribution}
\end{equation}
where $Z$ is a normalizing term; $T$ is a constant interpreted as the temperature in the statistical physics context and taken as equal to $1$ in this paper; $\frac{1}{Z}e^{-\frac{H(x)}{T}}$ is called the Gibbs distribution; and $H(x)$ is the potential energy of the system. 


The distribution function given by~(\ref{eq:gibbsdistribution}) enables us to compute the probability of the entire swarm reaching a certain configuration $\mathbf{x}$, but it requires global knowledge. 
As described in~\cite{rezeck2021flocking}, it is possible to obtain the following probability distribution that assumes only local information and enable us to sample velocities for each robot in a decentralized way:

\begin{equation}
    P_i(\mathbf{v_i}, \mathbf{\bar v_i}|\mathbf{x}) = 
    \frac{e^{-\left(U_{\{i\}}(\mathbf{\bar v_i}) + \sum \limits_{\forall j \in \mathcal{N}_i}U_{\{i,j\}}(\mathbf{\bar v_i}, \mathbf{v_j})\right)}}
    {\sum \limits_{\mathbf{z_i} \in \mathbf{Z}_i(x)} e^{-\left(U_{\{i\}}(\mathbf{z_i}) + \sum \limits_{\forall j \in \mathcal{N}_i}U_{\{i,j\}}(\mathbf{z_i}, \mathbf{v_j})\right)}},
    \label{eq:problocal}
\end{equation}
where $\mathbf{Z}_i(\mathbf{x}) \triangleq \{\mathbf{z}_i : ||\mathbf{z}_i|| \leq v_{max}\}$ is a set of possible velocities for the $i$-th robot given $\mathcal{N}_i$, restricted by  $\mathbf{Z}_i(\mathbf{x}) \subset \Lambda_i$; $U(\cdot)$ are potential functions; and $\mathbf{\bar v_i}$ is a velocity sampled in $\mathbf{Z}_i(\mathbf{x})$ representing a likely velocity for the next time step, $\displaystyle{\mathbf{v_i}^{(t+1)} = \mathbf{\bar v_i}}$.

Note that to sample over the distribution function~(\ref{eq:problocal}), it is required to define the potential energy as a combination of appropriate potential functions that quantify the state of the swarm concerning a physical property or behavior.

\subsection{Potential energy}
The potential energy $H(x)$ consists of the summation of values produced by potential functions, $U_A : \Lambda \rightarrow \mathbb{R}$. Formally, it is defined by pairwise interactions between neighboring vertices as
\begin{equation}
H(x) = \sum \limits_{i \in \mathcal{R}} U_{\{i\}}(\mathbf{v}_i) + \sum \limits_{(i,j) \in \mathcal{R}  \times \mathcal{R}, j \in \mathcal{B}_i} U_{\{i,j\}}(\mathbf{v}_i, \mathbf{v}_j),
\label{eq:potentialenergy2}
\end{equation}
where $U_{\{i\}}(\mathbf{v}_i)$ is interpreted as the potential for the $i$-th robot to reach the velocity $\mathbf{v}_i$, and $U_{\{i,j\}}(\mathbf{v}_i, \mathbf{v}_j)$ is the potential regarding the velocities of the neighboring $(\mathbf{v}_i, \mathbf{v}_j)$ pair of vertices.

Here we propose combining the Coulomb-Buckingham potential with the kinetic energy. The kinetic energy is computed from the relative velocities among all robots in the bond partition and induces the group to \rev{reach consensus} on their velocities \rev{allowing for} cohesive navigation. The kinetic energy $\mathbf{E_k}$ relative to the $i$-th robot is:
\begin{equation}
\mathbf{V}_i = \sum \limits_{\forall j \in \mathcal{B}_i} \mathbf{v}_j, \ \ \ \mathbf{E_k}(\mathbf{V}_i) = \frac{1}{2} m (\mathbf{V}_i \cdot \mathbf{V}_i),
\end{equation}
where $m$ is the cumulative mass of the group.

The Coulomb-Buckingham potential~\cite{buckingham1938classical} is a combination of the Lennard-Jones potential with the Coulomb potential used to describe the interaction among particles considering their charges. The formula for the interaction is
\begin{equation}
\Phi (r)=\varepsilon \left({\frac {6}{\alpha -6}}e^{\alpha} \left(1-{\frac {r}{r_{0}}}\right)-{\frac {\alpha }{\alpha -6}}\left({\frac {r_{0}}{r}}\right)^{6}\right) + \frac {c_{i}c_{j}}{4\pi \varepsilon_0 r},
    \label{eq:cbpotential}
\end{equation}
where $r = ||\mathbf{q}_j - \mathbf{q}_i||$ is the euclidean distance between the particles $i$ and $j$; $\varepsilon$ is the depth of the minimum energy; $r_{0}$ is the minimum energy distance; $\alpha$ is a free dimensionless parameter; $c_i$ and $c_j$ are the charges of the particles $i$ and $j$; and $\varepsilon_0$ is an electric constant. 

Considering the $i$-th robot, the interaction with $j$-th robot follows the constraints established by it bond partition $\mathcal{B}_i$, and the product $c_i c_j$ is defined by,
\begin{equation}
    C(i,j) = \left(1 - 2~\mathbb{1}(j \in \mathcal{B}_i ) \right)|c_i c_j|,
\end{equation}
where $\mathbb{1}(\cdot)$ indicates if the $j$-th is on the partition $\mathcal{B}_i$. If so, then $|c_i c_j|$ is set negative generating attractiveness among the $i$-th and $j$-th robos. Otherwise, they repel each other. 



The final potential energy term is computed combining Coulomb-Buckingham potential and the kinetic energy as described in~\cite{rezeck2021flocking}.

\subsection{Sampling algorithm}
Finally, given the probability function~(\ref{eq:problocal}) and the potential energy one may use a MCMC algorithm to sample velocities for the $i$-th robot. In this work, we use the Metropolis-Hastings algorithm~\cite{hastings1970monte}, one of the most widely used and efficient methods for sampling probabilities when the normalizing constant is unknown. As the conventional Metropolis-Hastings algorithm was originally proposed for discrete state spaces, we use an alternative method presented by Walker~\cite{walker2014sampling} for sampling on a countably infinite state space of velocities. This alternative to the Metropolis-Hastings algorithm is presented in  Algorithm~\ref{alg:metropolis}.

\setlength{\textfloatsep}{0.4cm}
\begin{algorithm}[!t]
\caption{Sampling velocity for $i$-th robot at time $t+1$}\label{alg:metropolis}
\begin{algorithmic}[1]
\Procedure{Update}{$\mathbf{q_i}^{(t)}, \mathbf{v_i}^{(t)}$} \Comment{\rev{Current state}}
 \State $\mathbf{V}^{(0)} \gets \mathbf{v_i}^{(t)}$, \Comment{Set of velocities} 
 \State $\mathbf{U}^{(0)} \gets H(\mathbf{q_i}^{(t)}, \mathbf{v_i}^{(t)})$, \Comment{Set of potential energies} 
 \For{$k \gets 1$ to $\mathcal{I}$}
 \State $\bar{\mathbf{v}} \gets N(\mathbf{v_i}^{(t)}, \mathbf{\Sigma})$,    \Comment{Gaussian sampling} 
 \State $\bar{\mathbf{u}} \gets  H(\mathbf{q_i}^{(t)}, \bar{\mathbf{v}})$,
 \State $\Delta E \gets \bar{\mathbf{u}} - \mathbf{U}^{(k-1)}$,      \Comment{Potential energy variation}
 \State $g \gets exp(-\Delta E)$,   \Comment{Compute the Gibbs energy}
 \State $r \gets \mathcal{U}(0, 1)$,   \Comment{Uniform sampling}
 \If{$(\Delta E < 0) \lor (r < g)$ } \Comment{Sample accepted}
    \State $\mathbf{V}^{(k)} \gets \bar{\mathbf{v}}$,
    \State $\mathbf{U}^{(k)} \gets \bar{\mathbf{u}}$,
    \Else \Comment{Sample rejected}
    \State $\mathbf{V}^{(k)} \gets  {\mathbf{V}^{(k-1)}}$,
    \State $\mathbf{U}^{(k)} \gets  {\mathbf{U}^{(k-1)}}$,
    \EndIf
 \EndFor
 \State $\mathbf{V} \gets \mathbf{V}^{(j,...,\mathcal{I})}$, \Comment{Reject the first j velocities}
 \State $\mathbf{v_i}^{(t+1)} \gets  (\mathbf{V}^q + ...+\mathbf{V}^\mathcal{I})/(\mathcal{I}-q)$, \Comment{Average}
 \State \textbf{return} $\mathbf{v_i}^{(t+1)}$
\EndProcedure
\end{algorithmic}
\end{algorithm}
\section{Experiments and Results}
\label{sec:experiments}
This section demonstrates how a swarm of heterogeneous robots can create diverse patterns using different neighborhood constraints. Further on, we use a realistic simulator as well as real-robot experiments to show that our method may serve as basis for more tangible applications, for example the construction of chain/bridge like structures. \rev{The source code and videos of each experiment are available online\footnote{\scriptsize \url{https://rezeck.github.io/gibbs_swarm_pattern_formation/}}}.

\subsection{Diversity of patterns}
To evaluate the versatility \rev{and efficacy} of our method in producing different types of patterns with a swarm of robots, we show through numerical simulations four examples where each one has different maximum bond constraints. In all examples, we assume a heterogeneous swarm of $\eta=180$~robots, uniformly distributed in $10 \times 10$~meters bounded environments. The robots are driven by a holonomic kinematic model, \rev{reaching a maximum speed of $v_{max}=1.0$~meters per second and have a maximum sensing range of $\lambda=0.5$~meters. We performed $100$ runs with a maximum of $20000$ iterations and analyzed the consensus speed among robots in the same group and the persistence of the patterns. As a metric for the persistence, we compute the number of remaining bonds and the number of molecules formed by the swarm over time. Here we consider a molecule as a group of robots that bonds together and has no remaining bonds.}

\begin{figure}[t]
\vspace{0.1cm}
    \begin{subfigure}{.50\textwidth}
		\centering
        \begin{minipage}[c]{0.03\textwidth}
            \caption{} \label{fig:pattern_water}
        \end{minipage}\hfill	
        \begin{minipage}[c]{0.96\textwidth}
            \includegraphics[width=.94\linewidth]{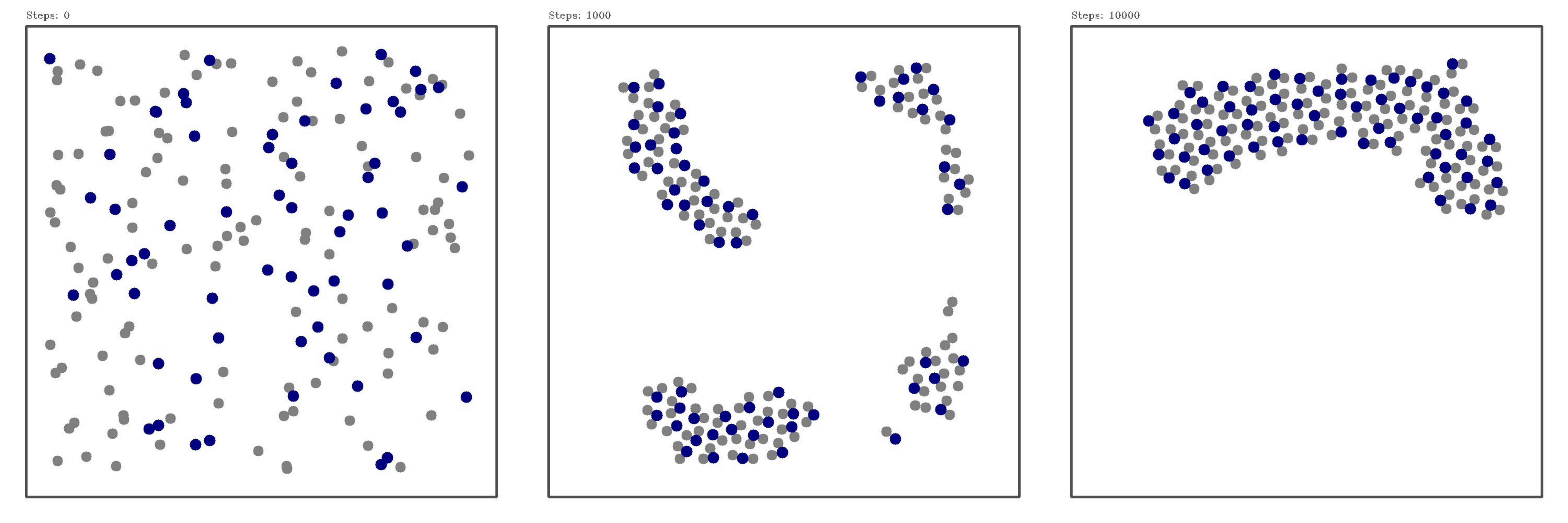}
        \end{minipage}
  	\end{subfigure}%
    \quad
    \begin{subfigure}{.50\textwidth}
		\centering
        \begin{minipage}[c]{0.03\textwidth}
            \caption{} \label{fig:pattern_methane}
        \end{minipage}\hfill	
        \begin{minipage}[c]{0.96\textwidth}
            \includegraphics[width=.94\linewidth]{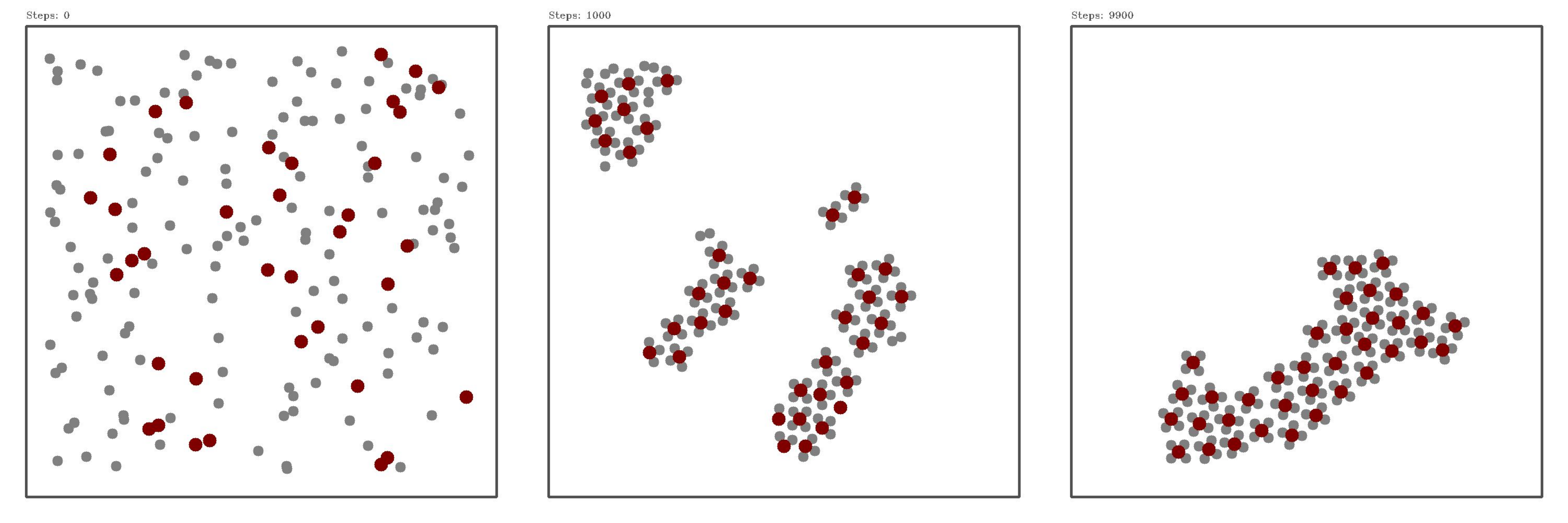}
        \end{minipage}
  	\end{subfigure}%
    \quad
    \begin{subfigure}{.50\textwidth}
		\centering
        \begin{minipage}[c]{0.03\textwidth}
            \caption{} \label{fig:pattern_polyamines}
        \end{minipage}\hfill	
        \begin{minipage}[c]{0.96\textwidth}
            \includegraphics[width=.94\linewidth]{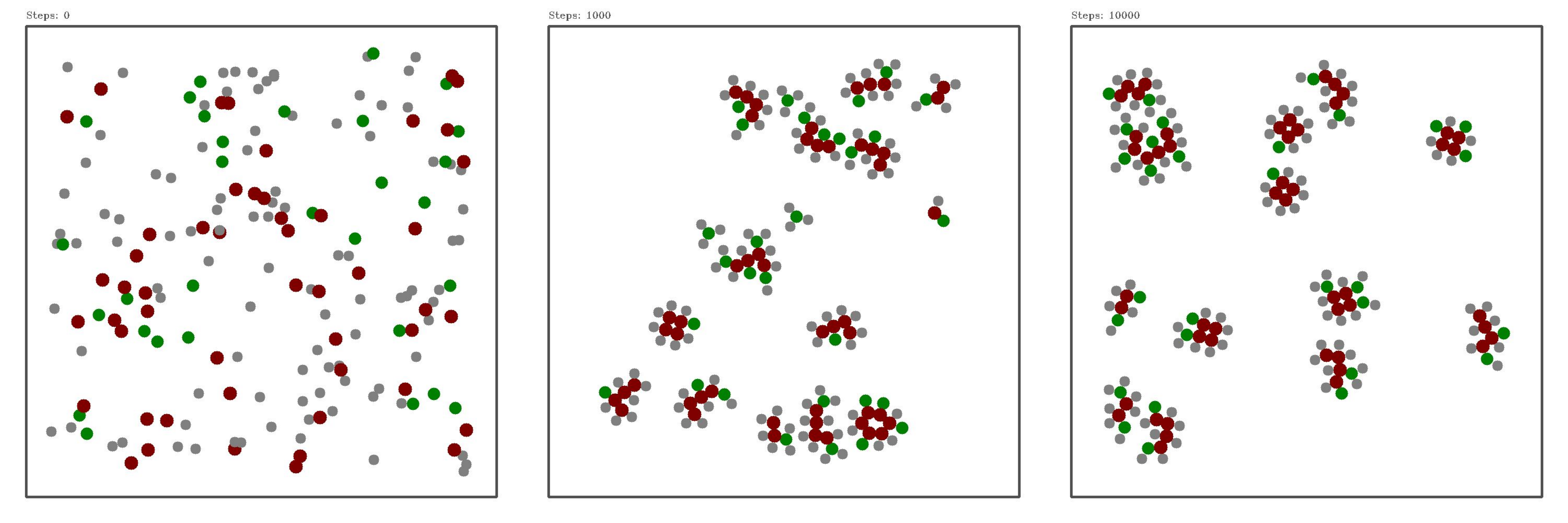}
        \end{minipage}
  	\end{subfigure}%
    \quad
    \begin{subfigure}{.50\textwidth}
		\centering
        \begin{minipage}[c]{0.03\textwidth}
            \caption{} \label{fig:pattern_oxocarbon}
        \end{minipage}\hfill	
        \begin{minipage}[c]{0.96\textwidth}
            \includegraphics[width=.94\linewidth]{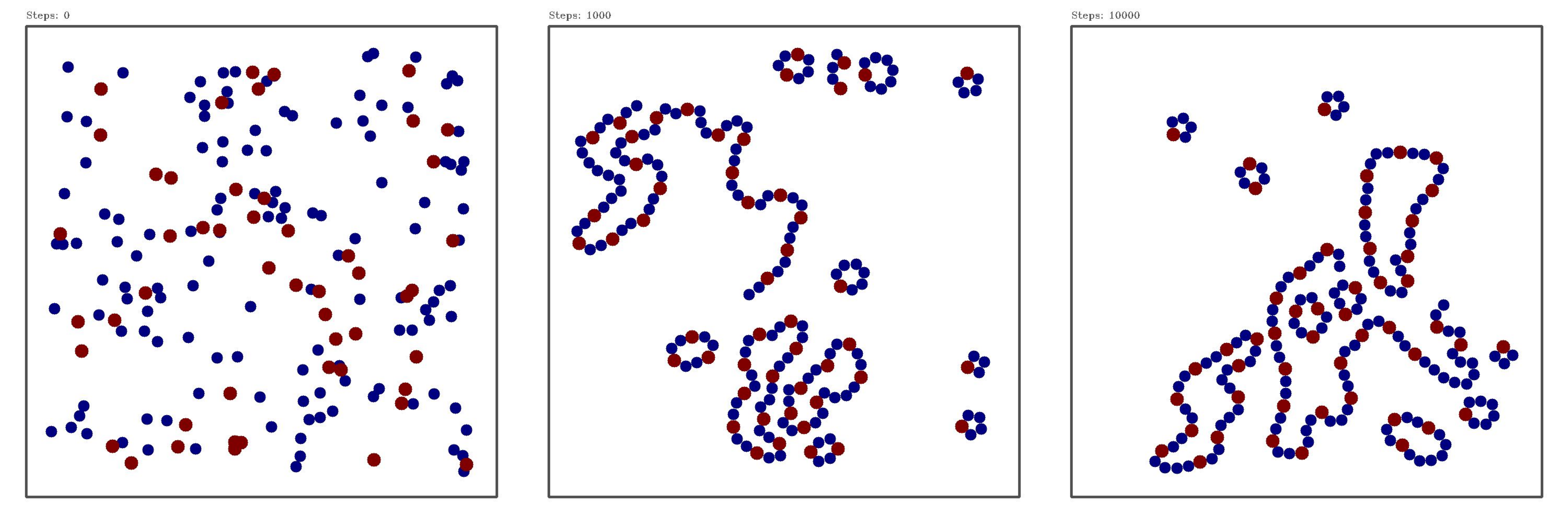}
        \end{minipage}
  	\end{subfigure}%
	\caption{Snapshots of four simulated experiments showing that different types of robots create different patterns resembling chemical structures of the (a) water, (b) methane, (c) polyamines and (d) oxocarbons.}\vspace{-0.2cm}
	\label{fig:pattern_snapshots}
\end{figure}

\begin{figure}[t]
\vspace{0.2cm}
    \begin{subfigure}{.50\textwidth}
		\centering
        \begin{minipage}[c]{0.03\textwidth}
            \caption{} \label{fig:analyses_water}
        \end{minipage}\hfill	
        \begin{minipage}[c]{0.96\textwidth}
            \includegraphics[width=.85\linewidth]{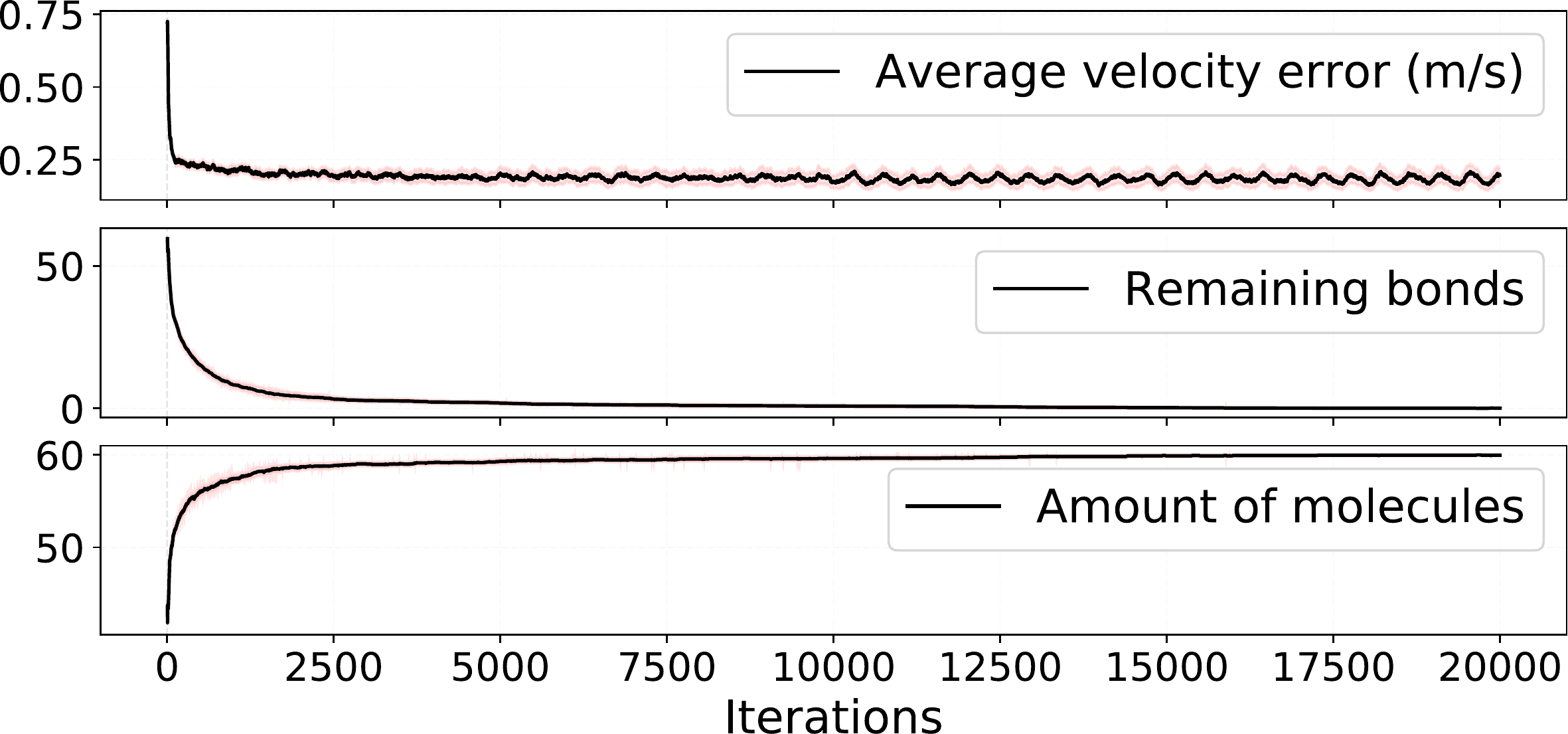}
        \end{minipage}
  	\end{subfigure}%
    \quad
    \begin{subfigure}{.50\textwidth}
		\centering
        \begin{minipage}[c]{0.03\textwidth}
            \caption{} \label{fig:analyses_methane}
        \end{minipage}\hfill	
        \begin{minipage}[c]{0.96\textwidth}
            \includegraphics[width=.85\linewidth]{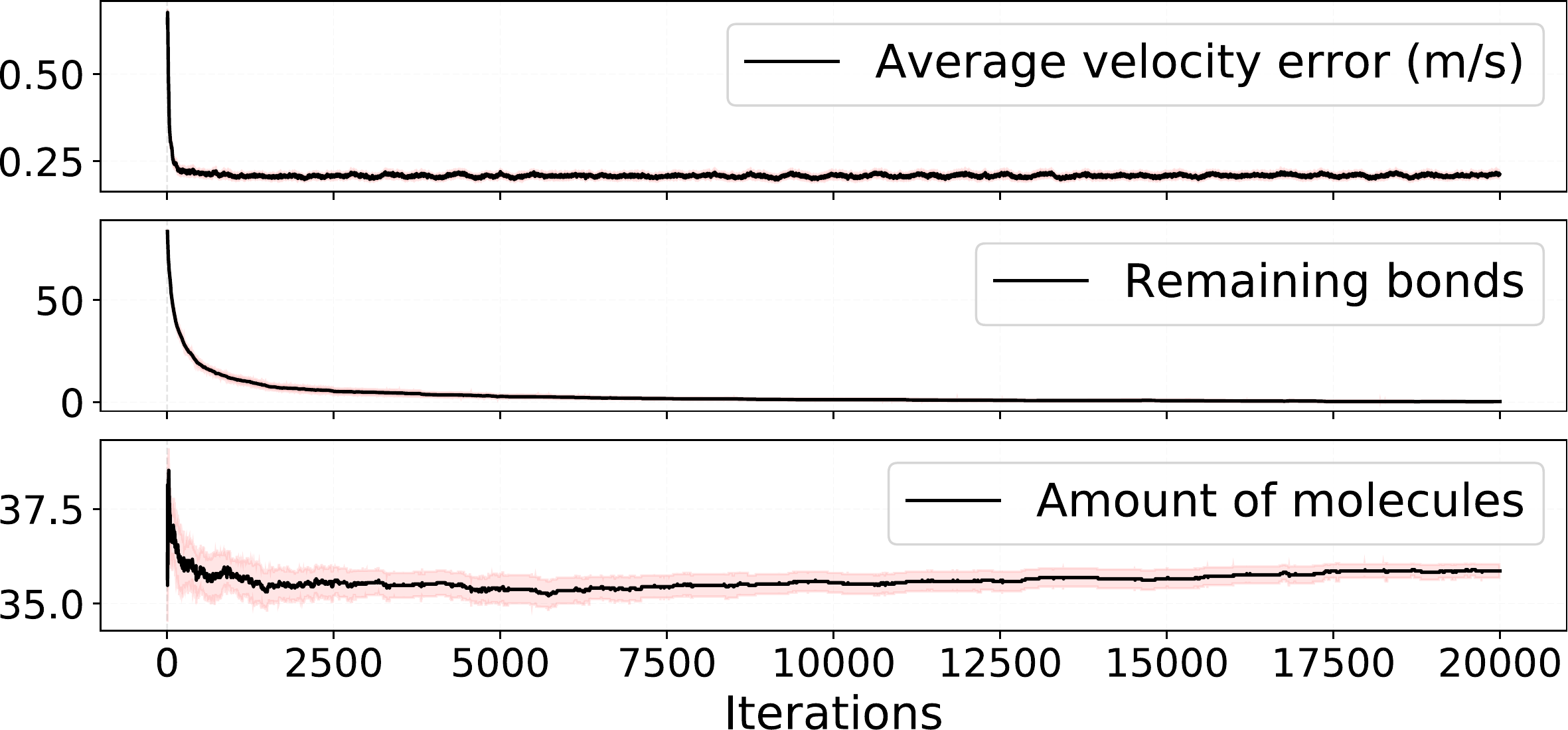}
        \end{minipage}
  	\end{subfigure}%
    \quad
    \begin{subfigure}{.50\textwidth}
		\centering
        \begin{minipage}[c]{0.03\textwidth}
            \caption{} \label{fig:analyses_polyamines}
        \end{minipage}\hfill	
        \begin{minipage}[c]{0.96\textwidth}
            \includegraphics[width=.85\linewidth]{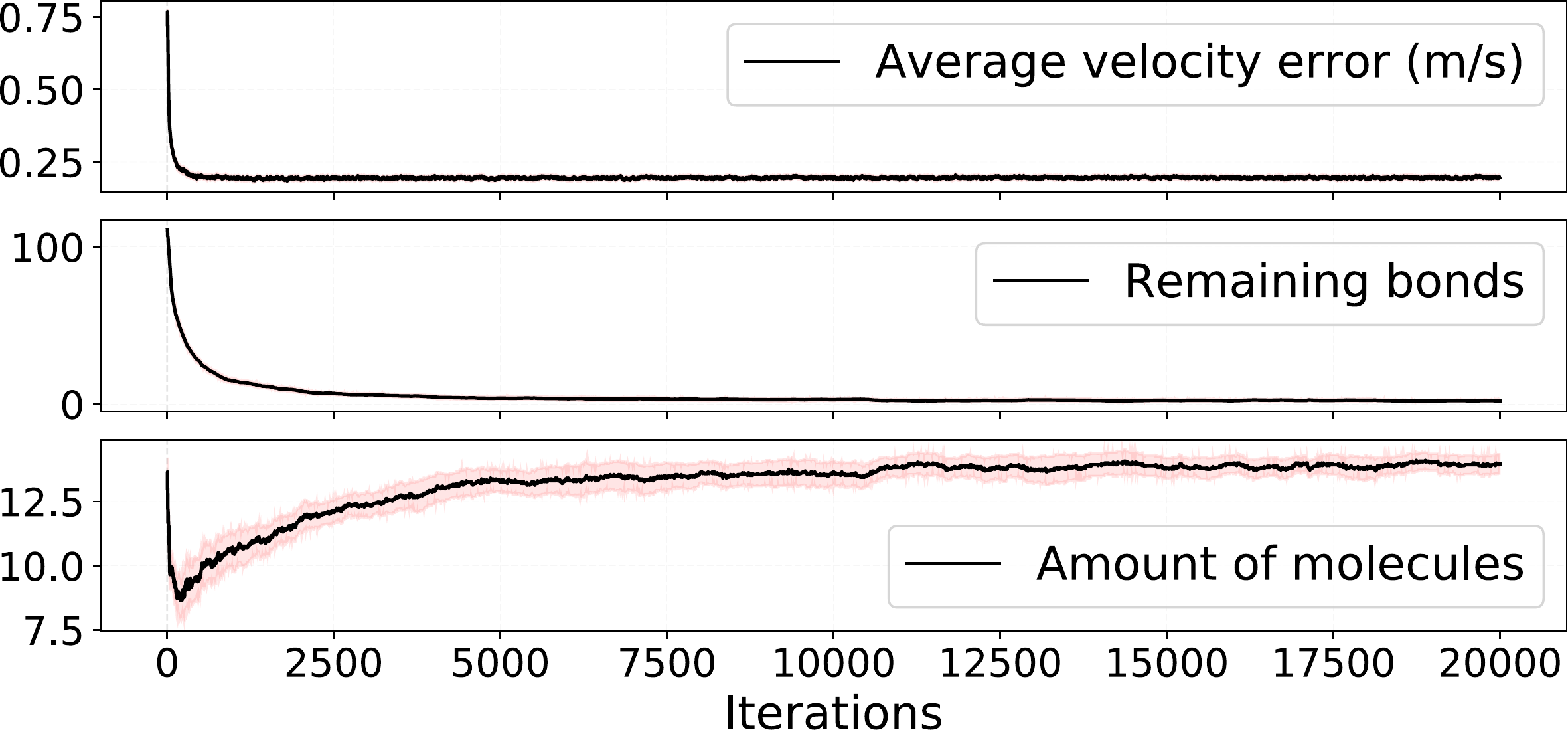}
        \end{minipage}
  	\end{subfigure}%
    \quad
    \begin{subfigure}{.50\textwidth}
		\centering
        \begin{minipage}[c]{0.03\textwidth}
            \caption{} \label{fig:analyses_oxocarbon}
        \end{minipage}\hfill	
        \begin{minipage}[c]{0.96\textwidth}
            \includegraphics[width=.85\linewidth]{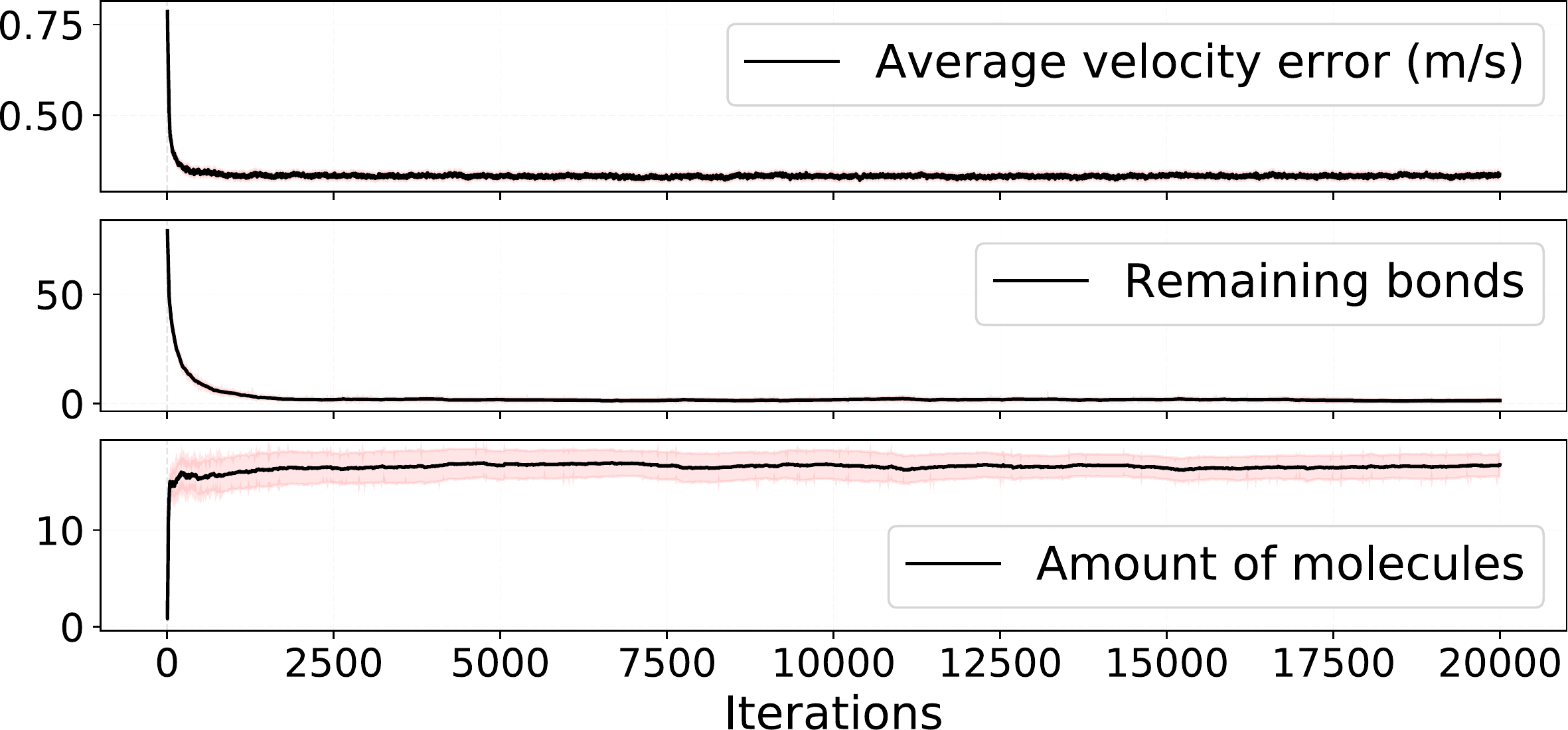}
        \end{minipage}
  	\end{subfigure}%
	\caption{\rev{Analysis of the persistence and velocity consensus for different patterns: (a) water, (b) methane, (c) polyamines, and (d) oxocarbons. The graphics show the mean and the $99\%$ confidence interval for $100$ runs measuring three different metrics. As metrics, we define the average velocity error for each group of robots, the number of remaining bonds, and molecules formed by the swarm in up to $20000$ iterations.}} \vspace{-0.3cm}
	\label{fig:pattern_snapshots}
\end{figure}

In the first example, we \rev{assume a swarm composed of} $|\tau| = 2$~types of robots: $|\tau_0|=120$  \textit{hydrogen}-like and $|\tau_1|=60$ \textit{oxygen}-like. The mass and electrical charge of the first one is $m_H=1$ and $||c_H||=1$, and the second one is $m_O=16$ and $||c_O||= 2$. We define its bond constraints as $(\mathcal{B}^\text{max}_{H,O}, \mathcal{B}^\text{max}_{H,H}) = (1, 1)$ and $(\mathcal{B}^\text{max}_{O,O}, \mathcal{B}^\text{max}_{O,H}) = (0, 2)$, respectively, and each type can have a maximum of $\mathcal{B}^\text{max}_{H} = 1$ and $\mathcal{B}^\text{max}_{O} = 2$ bonds \rev{allowing the swarm to form a maximum of $60$~molecules}. Figure~\ref{fig:pattern_water} shows a sequence of snapshots showing the swarm self-organizing to form structures that resemble water molecules. \rev{
Figure~\ref{fig:analyses_water} shows the mean and the $99\%$ confidence interval for each of the metrics. We note that the average velocity error in the swarm decreases and stabilizes at around $0.18 \pm 0.02$~meters per second. The mean error oscillations indicate cases where the swarm detects the borders of the environment, requiring the robots to change their velocity to avoid collisions. Regarding the creation and persistence of the patterns, we can see that the swarm converges to the desired patterns. All robots are bonded (only $0.07 \pm 0.05$ remaining bonds), and form the same number of groups (molecules), $59.96 \pm 0.09$, expected for this experiment.}

\rev{In the second example, we change the \textit{oxygen}-like robots for \textit{carbon}-like robots forming a swarm composed of $|\tau_0|=144$~\textit{hydrogen}-like and $|\tau_1|=36$ \textit{carbon}-like robots.} The mass and electrical charge of the \textit{carbon}-like robots are $m_C=12$ and $||c_C||=4$. We define the maximum bond constraints as $(\mathcal{B}^\text{max}_{H,C}, \mathcal{B}^\text{max}_{H,H}) = (1,1)$ and $(\mathcal{B}^\text{max}_{C,C}, \mathcal{B}^\text{max}_{C,H}) = (0,4)$, respectively, and each type allows a maximum of $\mathcal{B}^\text{max}_{H} = 1$ and $\mathcal{B}^\text{max}_{C} = 4$ bonds \rev{allowing the swarm to form a maximum of $36$~molecules}. Figure~\ref{fig:pattern_methane} shows the swarm building structures that resemble methane molecules. \rev{As seen in the Figure~\ref{fig:analyses_methane}, the swarm also reaches consensus in speed, stabilizing at around $0.21\pm0.01$~meters per second, and reduces the number of remaining bonds to $0.27\pm0.36$. Regarding the number of molecules, we observed a variation over the iterations but reaching an average of $35.86\pm0.18$ molecules. Unlike the previous pattern, the structures formed are more complex, and as there is a tendency for the molecules to aggregate, some robots may be blocked, making their bond challenging. However, as the method is dynamic, it eventually reaches the number of molecules expected for this experiment.}

\rev{In the third example, we use $|\tau| = 3$~types of robots: $|\tau_0|=120$ \textit{hydrogen}-like, $|\tau_1|=30$ \textit{nitrogen}-like and $|\tau_2|=30$ \textit{carbon}-like.} The mass and electrical charge of the \textit{nitrogen}-like robots are $m_N=14$ and $||c_N||=3$. We define maximum bond constraints as $(\mathcal{B}^\text{max}_{H,C}, \mathcal{B}^\text{max}_{H,N}, \mathcal{B}^\text{max}_{H,H}) = (1,1,1)$,  $(\mathcal{B}^\text{max}_{N,C}, \mathcal{B}^\text{max}_{N,N}, \mathcal{B}^\text{max}_{N,H}) = (1,0,3)$ and $(\mathcal{B}^\text{max}_{C,C}, \mathcal{B}^\text{max}_{C,N}, \mathcal{B}^\text{max}_{C,H}) = (2,1,2)$ , respectively, and each type can have a maximum of $\mathcal{B}^\text{max}_{H} = 1$, $\mathcal{B}^\text{max}_{N} = 3$ and $\mathcal{B}^\text{max}_{C} = 4$ bonds \rev{restricting the swarm to form a maximum of $15$~molecules}. Figure~\ref{fig:pattern_polyamines} shows the swarm forming structures that resemble chemical structure of polyamines. \rev{Figure~\ref{fig:analyses_polyamines} shows that the swarm managed to stabilize the average error in velocity (around $0.19\pm0.01$~meters per second) and also reduced the remaining amount of bonds to $2.20\pm0.80$. As this pattern has three types of robots, different molecules may occur, varying their quantity. However, we also see that this number tends to stabilize near to $13.96\pm0.38$, closer to the expected value.}

In the last example, we \rev{assume a swarm composed of $|\tau_0|=130$ \textit{oxygen}-like and $|\tau_0|=50$ \textit{carbon}-like robots}. We define maximum bond constraints as $(\mathcal{B}^\text{max}_{O,C}, \mathcal{B}^\text{max}_{O,O}) = (1,2)$ and $(\mathcal{B}^\text{max}_{C,C}, \mathcal{B}^\text{max}_{C,O}) = (0,2)$, respectively, and each type can have a maximum of $\mathcal{B}^\text{max}_{O} = 2$ and $\mathcal{B}^\text{max}_{C} = 2$ bonds \rev{restricting the swarm to form a maximum of $40$~molecules}. Figure~\ref{fig:pattern_oxocarbon} shows the swarm creating structures that resemble chemical structures of the oxocarbon. \rev{As with the previous patterns, we observed in Figure~\ref{fig:analyses_oxocarbon} that the swarm achieved consensus in velocity (around $0.33\pm0.01$~meters per second) and reduced the remaining bonds to $1.37\pm0.74$. 
We also observed a trend to converge to a specific number of molecules $16.75\pm1.15$, which is less than the maximum expected value. Unlike the other patterns, the rules allow the formation of larger molecules in the form of chains, reducing the total number of small molecules that the swarm can form.}

These experiments show that the swarm is capable of generating different patterns and dynamics depending on the binding constraints. In the first two, the swarm \rev{produced} sub-structures that \rev{ended} up aggregating. In the third experiment, we observed the formation of more complex and also varied structures. In the last experiment, the swarm \rev{created} long-chain forms with dynamic behavior that resemble biological systems. These are just a few examples of patterns that our approach can produce. 

However, although the method yields a variety of patterns, adjusting its restrictions so that the swarm forms desired structures, such as geometric shapes, is not straightforward without using some form of coordination mechanism. In this sense, in the next section we show how we place robot anchors to build and position chain like structures, which may be the keystone for more concrete applications.

\subsection{Building chain-like structures with real robots}
\label{Sec:RealExperiments}
While we take inspiration from chemical reactions so that a swarm can create different shapes in an emergent fashion, an important goal is to use  robot swarms for more practical applications. The chain-like oxocarbon structures created by our method are interesting candidates because of the directionality of the emergent patterns. \rev{In fact, structure in the form of chains has already been investigated to form paths for foraging problems~\cite{nouyan2008path, lee2022dynamic}. Here we consider employing} such behavior to create temporary bridges that could automatically adjust their size and shape to fit any building or terrain.

As a case study, we show through physical simulations and real-robot experiments the application of the method to dynamically form shapes with a topology similar to a bridge. For this, in addition to carbon-like and oxygen-like robots, we defined anchor robots to delimit the ends of the structure, allowing us to control the structure's using other algorithms or human-swarm interaction.

\begin{figure*}[h]
   \vspace{0.2cm}
	\begin{subfigure}{.20\textwidth}
		\centering
		\includegraphics[width=.90\linewidth]{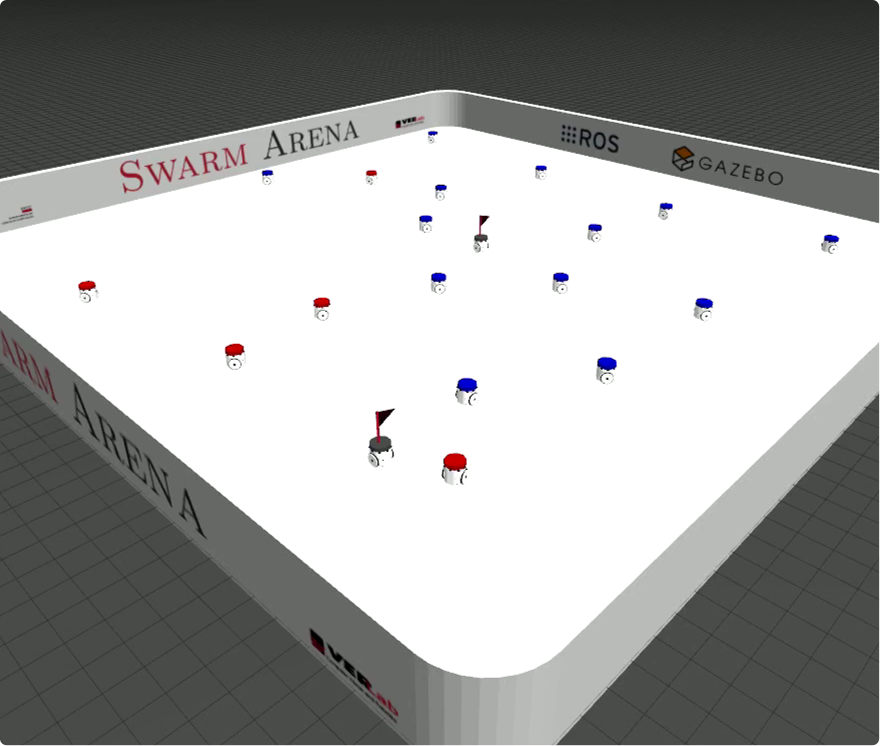}
		\caption{$t=0$~s.}
	\end{subfigure}%
    \hspace*{-0.5em}
	\begin{subfigure}{.20\textwidth}
		\centering
		\includegraphics[width=.90\linewidth]{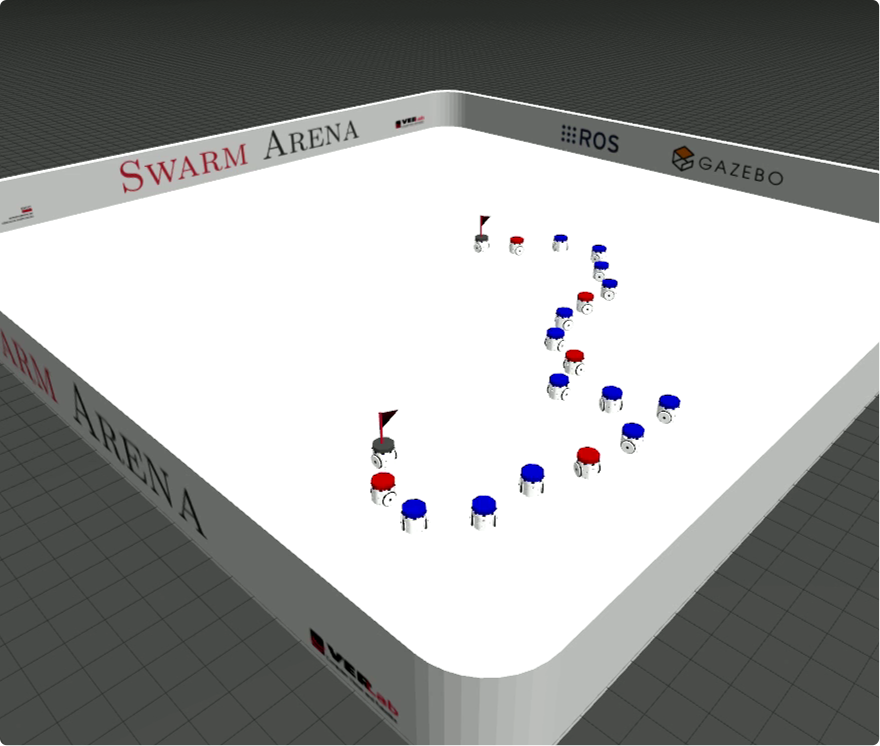}
		\caption{$t=115$~s.}
	\end{subfigure}%
    \hspace*{-0.5em}
	\begin{subfigure}{.20\textwidth}
		\centering
		\includegraphics[width=.90\linewidth]{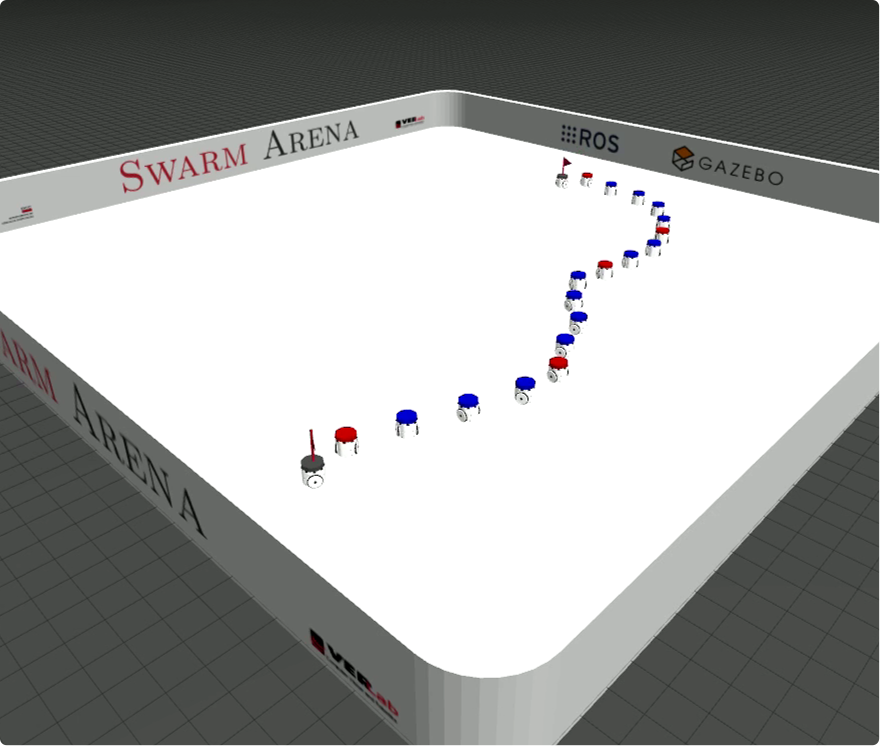}
		\caption{$t=135$~s.}
	\end{subfigure}%
	\hspace*{-0.5em}
	\begin{subfigure}{.20\textwidth}
		\centering
		\includegraphics[width=.90\linewidth]{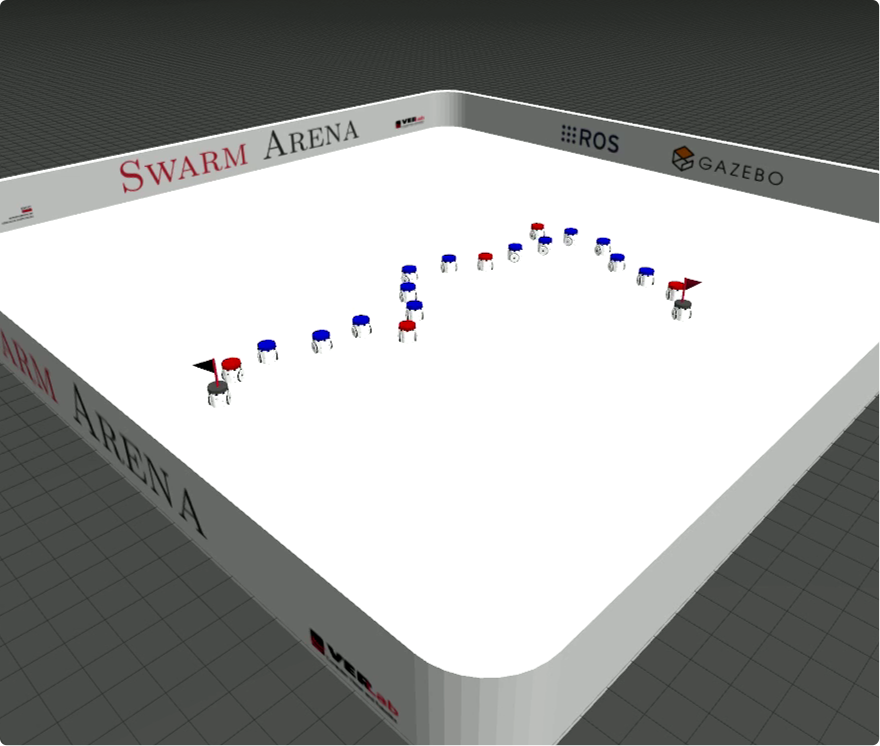}
		\caption{$n=158$~s.}
	\end{subfigure}%
	\hspace*{-0.5em}
	\begin{subfigure}{.20\textwidth}
		\centering
		\includegraphics[width=.90\linewidth]{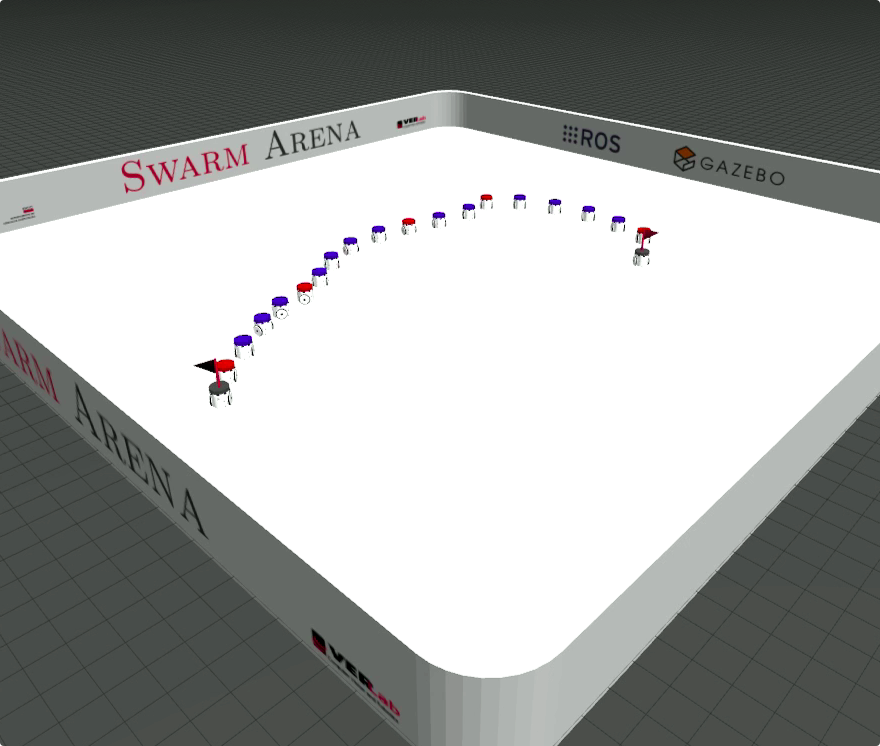}
		\caption{$n=170$~s.}
	\end{subfigure}%
	\caption{$20$ robots mimicking atoms of carbon (red) and oxygen (blue) to form shapes with a topology similar to a bridge. Robots with red flags delimit the ends of the structure. In the figure: (a) initial swarm configuration; (b) robots form the bridge; (c) change the position of robots with flag; (d) e (f) change again the position of robots with flag.} 
	\label{fig:bridge_gazebo}
\end{figure*}

We firstly set up an environment in Gazebo and implemented the methodology using the ROS (Robot Operating System) middleware. We simulated twenty HeRo Robots, a small and affordable differential-drive robot built in our laboratory~\cite{rezeck2017hero,rezeck2022hero}.
In this experiment, we distributed the robots around the environment and placed two static robots to delimit the beginning and end of the structure. Figure~\ref{fig:bridge_gazebo} shows the simulated results, where we can see the swarm converging on a structure that connects the two static robots. Also, when moving the anchor robots, we can see that the swarm reconfigures itself to maintain the structure.

\begin{figure*}[t] 
	\begin{subfigure}{.24\textwidth}
		\centering
		\includegraphics[width=.99\linewidth]{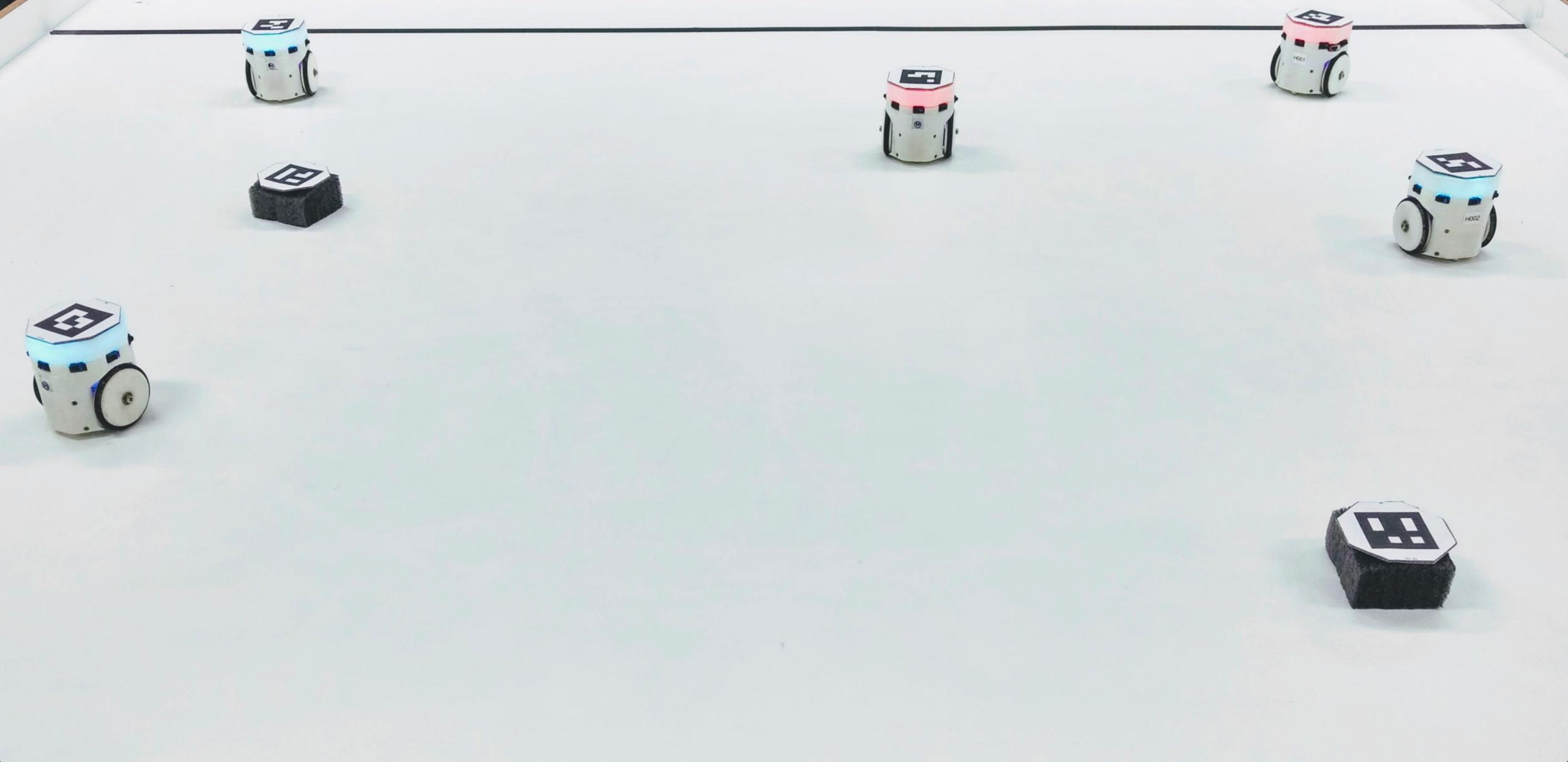}
		\caption{$n=0$~s.}
		\label{fig:real_a}
	\end{subfigure}
    \begin{subfigure}{.24\textwidth}
		\centering
		\includegraphics[width=.99\linewidth]{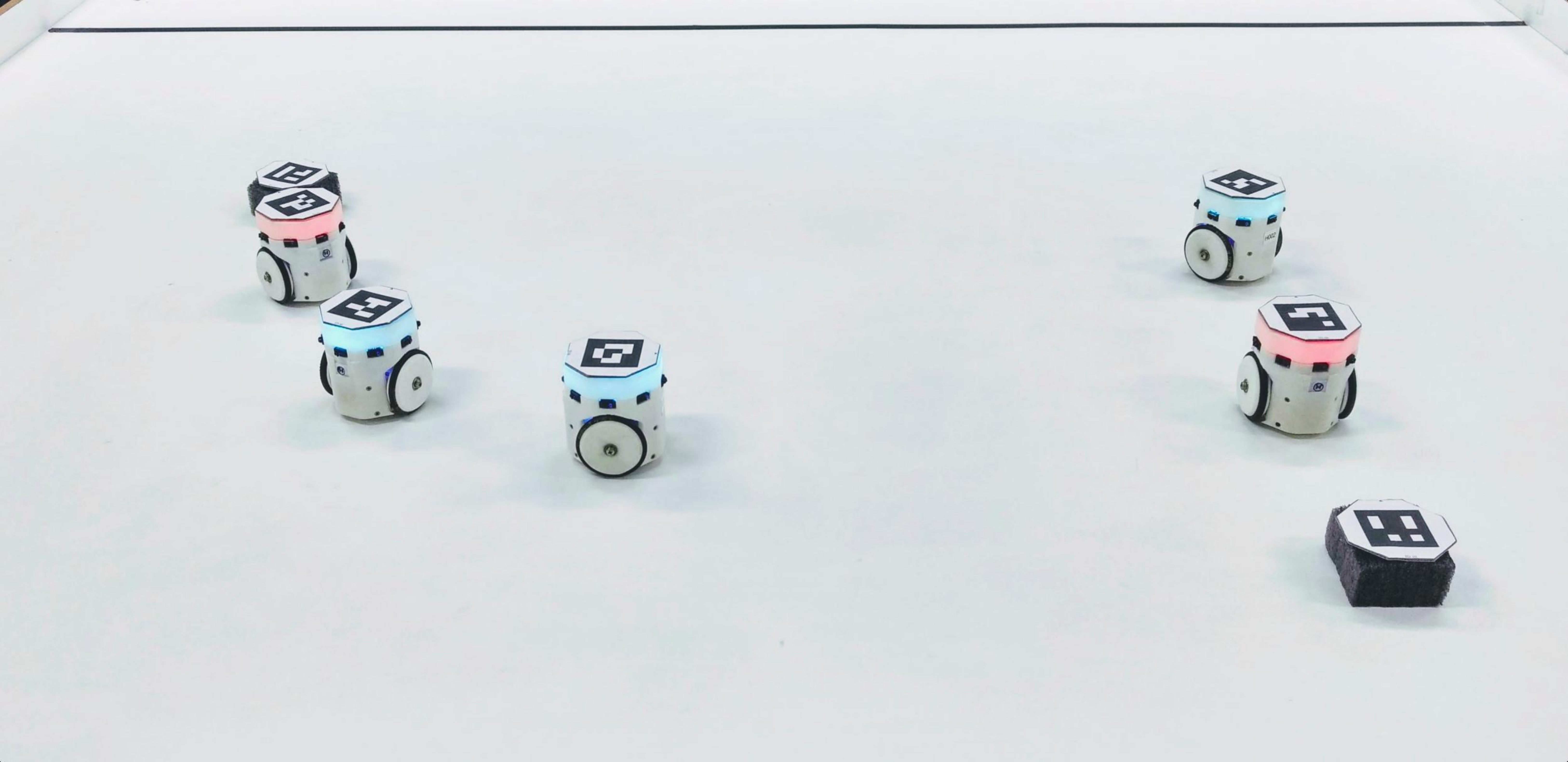}
		\caption{$n=30$~s.}
		\label{fig:real_a}
	\end{subfigure}
	\begin{subfigure}{.24\textwidth}
		\centering
		\includegraphics[width=.99\linewidth]{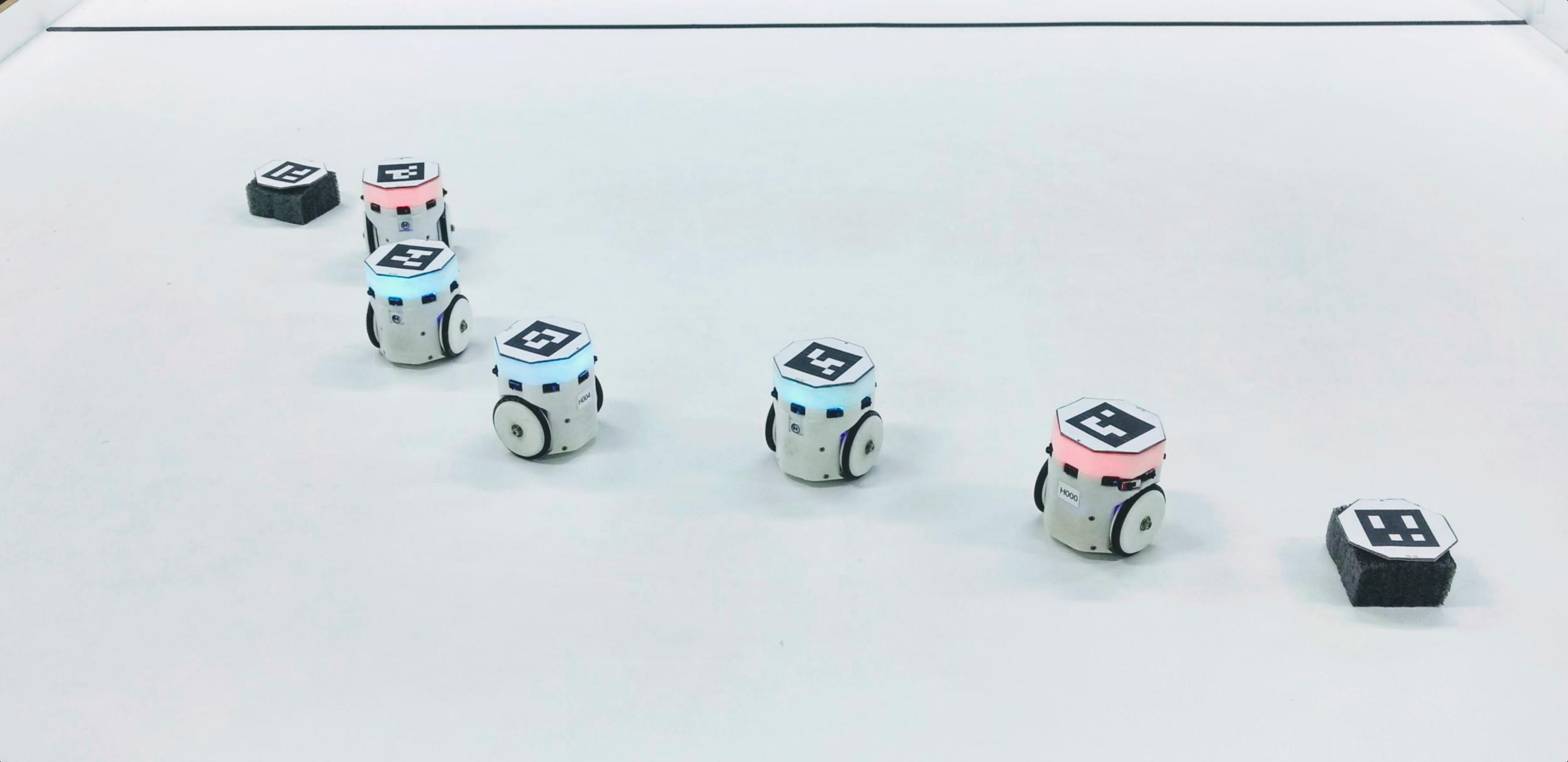}
		\caption{$n=50$~s.}
		\label{fig:real_a}
	\end{subfigure}
	\begin{subfigure}{.24\textwidth}
		\centering
		\includegraphics[width=.99\linewidth]{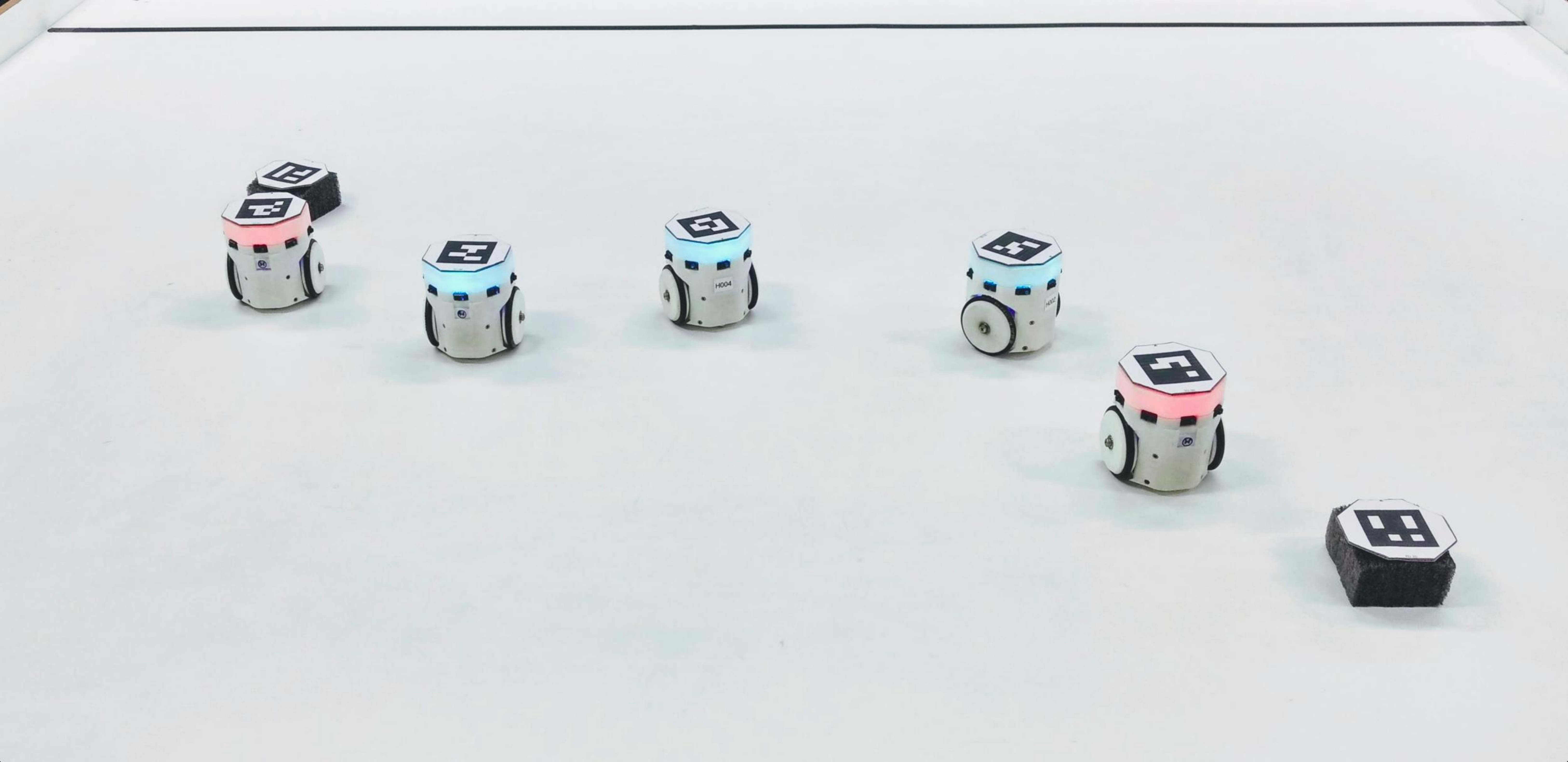}
		\caption{$n=70$~s.}
		\label{fig:real_a}
	\end{subfigure}
	\caption{Snapshots of an experiment with $5$ real robots mimicking $2$ carbon (red) and $3$ oxygen (blue) atoms creating a dynamic bridge pattern. Black blocks in the image delimit the beginning and end of the structure.}
	\label{fig:bridge_real}
\end{figure*}

We could also observe a similar behavior in proof-of-concept experiments with real robots, as shown in Figure~\ref{fig:bridge_real}. We used five HeRo robots remotely controlled by ROS and with local sensors emulated using images from an overhead camera. \rev{To control the robots to follow the velocity, we use the method described by Sordalen~\etal~\cite{sordalen1995feedback}.} The two black blocks in the image play the role of the anchor robots. The robots are able to form a chain-like structure and dynamically adapt its shape as we move the anchors.



\section{Conclusion and Future Work}
\label{sec:conclusion}

This paper presented a novel decentralized approach that allows a swarm of heterogeneous robots to form different patterns and shapes in an emergent fashion. Inspired by chemistry's Octet rule, we developed a mechanism to determine possible bindings among robots and improved our stochastic methodology based on Gibbs Random Fields to control the swarm. Experiments using numerical simulations demonstrated the versatility of our approach in producing different patterns by setting only the maximum bond constraints. Moreover, we presented physical simulations and real-robot experiments in which a swarm was capable of building a chain-like structure, which is potentially attractive for more tangible applications. 

\rev{We believe that our methodology has potential use in various scenarios where we want to build more complex structures from simple ones. One possible application is modular robotics, in which we make complex robots from simpler modules and can dynamically change their shape/structure. The bonding behaviors would guide the connection of these modules in a simple and dynamic fashion. Another application, as exemplified in our last experiment, is the construction of temporary structures such as bridges, pallets, and platforms, with different industry and military uses.}

\rev{As future work, we intend to tackle the problem of which compounds are required to generate a given shape. Finding such a correlation for producing complex shapes is not straightforward, so we plan to study strategies to define the binding constraints using reinforcement learning and dynamically change these constraints during execution to produce specific shapes. We also want to investigate the possible use of human-swarm interaction to allow the robots to dynamically create shapes on demand, which may be useful in different applications.}

\bibliographystyle{unsrt}
\bibliography{root}

\end{document}